\def\year{2020}\relax
\documentclass[letterpaper]{article} 
\usepackage{aaai20}  
\usepackage{times}  
\usepackage{helvet}  
\usepackage{courier}  
\usepackage[hyphens]{url}
\usepackage{graphicx}  
\usepackage{graphicx}
\usepackage{nicefrac}
\usepackage{booktabs}
\usepackage{multirow}
\usepackage{algorithm,algorithmicx,algpseudocode}

\usepackage{amsmath}
\usepackage{amssymb}

\newtheorem{mytheorem}{Theorem}

\newcommand{\myproof}{\noindent {\bf Proof:\ \ }}
\newcommand{\myqed}{\mbox{$\diamond$}}
\newcommand{\myOmit}[1]{}

\newcommand{\myosd}{\textsc{On\-li\-ne Se\-ri\-al Di\-c\-ta\-to\-r}}
\newcommand{\myorp}{\textsc{On\-li\-ne R\-an\-d\-om Pr\-io\-ri\-ty}}
\newcommand{\mymax}{\textsc{Ma\-x\-im\-um Li\-ke}}
\newcommand{\mypar}{\textsc{Pa\-re\-to Li\-ke}}
\newcommand{\mylike}{\textsc{Li\-ke}}
\newcommand{\myblike}{\textsc{Ba\-la\-n\-ced Li\-ke}}
\newcommand{\mycheckmark}{\mbox{\checkmark}}

\begin{document}
\title{Online Fair Division: A Survey}
\author{Martin Aleksandrov and Toby Walsh\\
Technical University Berlin and UNSW Sydney\\
Germany and Australia\\
martin.aleksandrov@tu-berlin.de, tw@cse.unsw.edu.au}

\maketitle
\begin{abstract}
We survey a burgeoning and promising new research 
area that considers the online nature of 
many practical fair division problems. We 
identify wide variety of such online fair division problems, 
as well as discuss new mechanisms and 
normative properties that apply to this online setting. 
The online nature of such fair division problems provides both 
opportunities and challenges such as the possibility
to develop new online mechanisms
as well as the difficulty of dealing 
with an uncertain future. 
\end{abstract}

\section{Introduction}

Fair division \cite{cakecut} is an important problem facing society today as
increasing economical, environmental, and other pressures require us
to try to do more with limited resources. 
Much previous work in fair division assumes the problem is offline
and fixed. That is, we suppose that the agents being allocated resources, and the resources
being allocated to these agents are all known and fixed.
But practical reality is often quite different \cite{wki2014,waaai2015}. Fair
division problems are often online, with either the agents, or 
the resources to be allocated, or both not being fixed
and potentially changing over time. 
This presents a number
of technical challenges and opportunities. 

Consider allocating deceased organs to patients, donated food to charities, electric vehicles to charging stations, viewing slots to a telescope, etc. We often cannot wait till all resources are available, preferences known or agents present before starting to allocate the
resources. For example, when a kidney is donated, it must be allocated
to a patient within a few hours. As a second example, when food items 
arrive at a food bank, they must be allocated to charities promptly. 
As a third example,
when allocating charging slots to electric cars, we may
not know when or where cars will arrive for charging in the future
but must commit to providing charging slots to cars now. 
As a fourth example, when allocating landing slots at an airport for
the next hour, there might be considerable uncertainty about 
the demands in six hours time. As a fifth example,
when scheduling viewing slots on an expensive telescope, 
we might need to commit to the current period, 
before future viewing conditions and demands are fully known. 

The {\em online} nature of such fair division problems
changes {\em the mechanisms} available to 
allocate items. For example, consider the well-known and widely used \emph{sequential allocation} mechanism
studied in \cite{indivisible,knwxaaai13,knwijcai13,awxijcai2015,akwxecai16,waaai16}.
This mechanism has agents taking turns to pick
their most preferred remaining item. 
This is used in many offline settings due to 
its simplicity and nice normative properties. For instance,
it can be used to return allocations where an agent's envy for the allocation of another agent is limited to at most one item. 
However, this mechanism cannot be used in an online setting as
an agent's most preferred item may
not be currently (or even ever) available. We, therefore, need
to develop new {\em online} mechanisms that account for the fact that some items might never become available.

The {\em online} nature of such fair division problems
also changes {\em the normative properties} that we might look
to demand. For example,  suppose we are interested in \emph{strategy-proof} mechanisms
that encourage agents to report their sincere preferences. 
In deciding if agents have any incentive
to misreport preferences in an online
setting, we may take into account the online nature
of the problem, and thus consider fixed past decisions and unknown future. This leads to
an {\em online} and weaker form of strategy-proofness.
It is weaker because agents have committed to
their past decisions and uncertainty about
the future reduces their strategic options. On the plus side,
this means that
it may be easier to achieve strategy-proofness in the
online setting than in the offline setting. 
On the minus side, this also means that
it could be harder in the online
setting to achieve other properties. 
For instance, items that arrive in the future may
require us to change the allocation of items 
made in the present to ensure Pareto efficiency.

\section{Dimensions of online fair division}
 
We can characterize models of problems concerning
online fair division along
a number of orthogonal dimensions.

\subsection{Resource types}

We can classify online fair division problems according to whether resources are divisible or indivisible. 
For instance, \cite{wadt11} considers 
an online cake cutting problem in which 
the resource (``the cake'') is divisible,
whilst \cite{aagwijcai2015,awijcai17,mswijcai17,awki17,awki17b} consider 
fair division problems where items arrive
in an online fashion but are indivisible. 

Further, we can characterize
online fair division problems along the number
and nature of those resources.
Do we have one or multiple resources,
and are those resources homogeneous or
heterogeneous? 
For instance, \cite{wadt11} considers 
a single, heterogeneous and divisible cake, 
whilst \cite{leftbehind} considers
multiple, homogeneous and divisible resources
(e.g.\ disk space, memory and CPUs in a server
farm). 

And, online fair division problems can be characterized on 
the number of resources allocated
to each agent. Do we allocate 
a single resource to each agent (e.g.\ a 
single organ to each patient in \cite{mswijcai2017,mswaies18})? 
Or do we allocate multiple items to 
each agent (e.g.\ multiple food products in 
\cite{aagwijcai2015}, multiple charging slots in 
\cite{gdagmmwijcai2019})? 

\subsection{Online features}

Another dimension on which to decompose online fair division problems is whether
agents, resources or both are online. 
For instance, \cite{wadt11} considers 
the resource (``the cake'') to be fixed, and the agents being allocated this resource
to arrive and depart over time. 
Think of people turning up over time for
a birthday party and cutting off slices of the cake for 
people as they arrive. 
On the other hand, \cite{aagwijcai2015} considers
the agents to be fixed, but the items 
to arrive over time and to be allocated immediately on arrival. 
Think of foods products being donated to a food bank over
the course of the day but 
being allocated immediately to charities on donation. 
And, in the third case, 
both agents and items arrive over time. 
Think of an online organ matching 
problem (e.g.\ \cite{mswaies18}) in which
both the patients and the organs can arrive
at particular points in time and, in addition, patients can also depart. 

\subsection{Mechanism features}

Another dimension to consider is whether mechanisms are 
informed or uninformed about the future. 
In the informed setting (e.g.\ 
\cite{procacciaonlinepast}), 
the mechanism might have information
about items yet to arrive. 
In the uninformed setting, such
information is not known. 
In this case, we may make an adversarial
risk-averse assumption that agents
act supposing the worst possible
future \cite{wadt11}. 
We can also distinguish between centralized and 
distributed mechanisms. 
Finally, we can distinguish between mechanisms
that make decisions one by one (e.g.\ \cite{aagwijcai2015})  
or in batches (e.g.\ \cite{procacciaonlineenvy}). 

\section{Formal background}

We suppose there are $n$ agents being allocated $m$ 
items. Allocations may be whole (e.g.\ in the case of 
indivisible items), fractional (e.g.\ in the case of
divisible items), or randomized (e.g.\ in the case of 
indivisible items, this represents a probability
distribution over whole allocations). 
Often, as in many other areas, 
we will assume additive utilities. 
Additive utilities offer a compromise between simplicity and
expressivity.
However, some work in this area has considered more
general utilities (e.g.\ \cite{awki2019} studies the more general
class of monotone utilities). 
We consider a number of classical normative
properties such as Pareto efficiency
and envy-freeness.
An allocation is \emph{Pareto efficient} iff there is no other
allocation where all agents have as much utility,
and at least one agent has strictly more. 
An allocation is \emph{envy-free} iff no agent has strictly 
greater utility for the items allocated to another agent than the utility for their own items. Envy-freeness is a desirable but often unachievable fairness property
(consider two agents and one indivisible item that they
both like). Therefore, we also consider a relaxation that
can be always achieved: \emph{envy-free up to one item} (EF1). An allocation
is EF1 iff no agent has envy for another agent's bundle, supposing we can remove
one item from this bundle. EF1 can 
be achieved with a simple round-robin mechanism that 
allocates the most desired item left to the next
agent in a round-robin fashion. 
We say that a mechanism is Pareto efficient / envy-free / EF1 iff
it only returns allocations that are Pareto efficient / envy-free /
EF1. See \cite{awpricai2019} for more details. 

\section{Technical approaches}

There are a number of technical responses to the 
challenges and opportunities introduced by the online 
nature of such fair division problems. 

\subsection{Online mechanisms}

One approach to deal with the challenges of online fair
division is to propose new mechanisms that exploit
the online nature of the fair division problem.
Indeed, many offline mechanisms cannot be used
in an online setting because they make assumptions, 
such as all the items are available at one time, that 
are violated. We, therefore, often need to ``flip'' mechanisms
around to work in an online fashion.

Many mechanisms for the offline fair division 
of indivisible items take an agent-centered view
where one or more agents get to pick an item or items from a set of items. 
For example, in offline fair division, the probabilistic 
serial mechanism \cite{probserial} has 
every agent ``eating'' their most preferred item
at uniform speed, generating a randomized
allocation of items to agents. 
Such offline mechanisms need to be adapted
to work in an online setting as an agent's 
most preferred item might not be currently, or even ever
available. 

An online setting where items arrive over time
also naturally invites an item-centric view. 
For example, \cite{aagwijcai2015} propose
the \mylike\ mechanism where
an item on arrival is allocated 
uniformly at random between agents
that declare strictly positive utility
for the item. This online mechanism
has nice normative properties. 
For example, it is
strategy-proof and envy-free in expectation.
The \mylike\ mechanism can be seen as the online analog of the
(offline) {probabilistic serial} mechanism
with agents ``eating'' each next item which they like.

Many other online mechanisms for the fair division of indivisible items can 
be described as choosing uniformly between some subset of agents that are feasible
for an arriving item. Thus, an agent from this set receives the
item with \emph{conditional probability} that is uniform with respect
to the other agents that are feasible for
the item. Such mechanisms return a probability distribution
over allocations,
and an actual allocation with some positive probability that is
obtained as a product of 
the conditional randomizations. Some such mechanisms are studied in \cite{awpricai2019}:

\begin{itemize}
\item \myosd: there is a strict priority order $\sigma$ of the agents
  prior to round one, and the unique feasible agent for each next item is the first agent in $\sigma$ that bids positively for it.

\item \myorp: this draws uniformly at random a strict priority order $\sigma$ of the agents prior to round one, and runs \myosd\ with it. 
  
\item \mypar: an agent is feasible for an item if allocating the item
  to the agent is Pareto efficient ex post. 

\item \mylike: an agent is feasible for an item if they bid positively
  for it. 

\item \myblike: an agent is feasible for an item if they bid positively
  for it, and they are amongst the agents bidding positively with the
  fewest
items in the current allocation. 

\item \mymax: an agent is feasible for an item if they are amongst the
  agents making the largest bid for it. 
\end{itemize}

It should be clear that the online setting introduces a new design
space in which to define mechanisms for fair division. It is likely
that there are many interesting parts of this space yet to explore. 

\subsection{Online properties}

Another technique to deal with the challenges of online fair
division is to consider relaxed normative properties
that are applicable to the online setting. 
For example, \cite{awpricai2019} relax the definition of strategy-proofness
to suppose past decisions are fixed while future decisions could 
still be strategic. 
A mechanism is {\bf strategy-proof} if, an agent 
cannot improve their (expected) utility
by bidding insincerely, supposing the entire information (i.e.\ the items, their arriving order and the mechanism) about the allocation process is available to each agent. 
That is, for each problem and agent
$a_i$, $\overline{u}(u_{i1},\ldots,u_{im}) \geq
  \overline{u}(v_{i1},\ldots,v_{im})$
for any (possibly strategic) bids $v_{i1}, \ldots,v_{im}$
of agent $a_i$ for the $m$ items. 
On the other hand, we can relax this definition
to a weaker online version of strategy-proofness. 
We say that a mechanism is {\bf online strategy-proof} if, for each
problem and round $j \in [1,m]$,
$\overline{u}(u_{i1},\ldots,u_{i(j-1)},u_{ij}) \geq
  \overline{u}(u_{i1},\ldots,u_{i(j-1)},v_{ij})$
for any (possibly strategic) bid $v_{ij}$ of agent
$a_i$ for the final $j$th item. 

Online strategy-proofness is less onerous to achieve than
strategy proofness. It supposes all earlier decisions are now
fixed and the agent can only be strategic about the current decision. 
It is not hard to see that
a strategy-proof mechanism is further online strategy-proof
but the opposite may not hold. Indeed, whilst the only online mechanisms
that are strategy-proof are rather unresponsive mechanisms
such as the random mechanism which just allocates
items randomly irrespective of the bids of agents, the class of mechanisms
that are online strategy-proof is far larger and can be responsive
to the bids of agents. In fact, Theorem 2 in \cite{awpricai2019} characterizes the class of online mechanisms that are online
strategy-proof as ``step'' mechanisms that
allocate items uniformly between agents
that bid over some threshold of utility for an
arriving item. 

Stronger normative properties can also be useful in 
the online setting. 
For instance, \cite{awpricai2019} 
introduce {\bf shared envy-freeness} which
requires that each pair of agents are envy-free of each other 
over only the items that both agents in the pair like in common.
If you get an item that I don't like, it does not cause
me any envy. 
Shared envy-freeness implies envy-freeness but not the other way around. Shared envy-freeness is one of the normative
properties that distinguishes between the \mylike\ and
\myblike\ mechanisms. Table~\ref{tab:results}, which is taken from
\cite{awpricai2019}, summarizes many of the differences
in normative properties of the different online
mechanisms discussed in the previous section. 

\begin{table*}[htb]
\centering

\resizebox{0.75\textwidth}{!}{
\begin{tabular}{|c|c|c|c|c|c|c|c|c|c|}
\hline
\multirow{2}{*}{ {\bf Mechanism}} &  SP &  OSP &  EFA &  SEFA &  EFP &  SEFP &  BEFP &  PEA &  PEP \\ \cline{2-10}
& \multicolumn{9}{c|}{ general cardinal utilities} \\ \hline

 {\sc Online Random Priority} &  $\mycheckmark$ &  $\mycheckmark$ &  $\mycheckmark$ &  $\mycheckmark$ &  $\times$ &  $\times$ &  $\times$ &  $\times$ &  $\mycheckmark$ \\ \hline
 {\sc Online Serial Dictator} &  $\mycheckmark$ &  $\mycheckmark$ &  $\times$ &  $\times$ &  $\times$ &  $\times$ &  $\times$ &  $\mycheckmark$ &  $\mycheckmark$ \\ \hline
 {\sc Maximum Like} &  $\times$ &  $\times$ &  $\times$ &  $\times$ &  $\times$ &  $\times$ &  $\times$ &  $\mycheckmark$ &  $\mycheckmark$ \\ \hline
 {\sc Pareto Like} &  $\times$ &  $\times$ &  $\times$ &  $\times$ &  $\times$ &  $\times$ &  $\times$ &  $\times$ &  $\mycheckmark$ \\ \hline
 {\sc Like} &  $\mycheckmark$ &   $\mycheckmark$ &  $\mycheckmark$ &  $\mycheckmark$ &  $\times$ &  $\times$ &  $\times$ &  $\times$ &  $\times$ \\ \hline
 {\sc Balanced Like} &  $\times$ &  $\mycheckmark$ &  $\times$ &  $\times$ &  $\times$ &  $\times$ &  $\times$ &  $\times$ &  $\times$ \\ \hline
& \multicolumn{9}{c|}{ identical cardinal utilities} \\ \hline
 {\sc Like} &  $\mycheckmark$ &  $\mycheckmark$ &  $\mycheckmark$ &  $\mycheckmark$ &  $\times$ &  $\times$ &  $\times$ &  $\mycheckmark$ &  $\mycheckmark$ \\ \hline
 {\sc Balanced Like} &  $\times$ &  $\mycheckmark$ &  $\mycheckmark$ &  $\mycheckmark$ &  $\times$ &  $\times$ &  $\times$ &  $\mycheckmark$ &  $\mycheckmark$ \\ \hline

& \multicolumn{9}{c|}{ binary cardinal utilities} \\ \hline
 {\sc Like} &  $\mycheckmark$ &  $\mycheckmark$ &  $\mycheckmark$ &  $\mycheckmark$ &  $\times$ &  $\times$ &  $\times$ &  $\mycheckmark$ &  $\mycheckmark$ \\ \hline
 {\sc Balanced Like} &  $\times$ &  $\mycheckmark$ &  $\mycheckmark$ &  $\times$ &  $\times$ &  $\times$ &  $\mycheckmark$ &  $\mycheckmark$ &  $\mycheckmark$ \\ \hline
\end{tabular}
}
\caption{Normative properties of different online mechanisms. 
SP is strategy proof. 
OSP is online strategy proof.
EFA is envy-free ex ante.
SEFA is shared envy-free ex ante.
EFP is envy-free ex post.
SEFP is shared envy-free ex post.
BEFP is bounded envy-free ex post.
PEA is Pareto efficient ex ante.
PEP is Pareto efficient ex post.
}
\label{tab:results}
\end{table*}

\subsection{Impossibility results}

Another approach to understand the challenges of online fair
division is to identify normative properties which are
impossible to achieve in the online setting which
can be achieved in the offline setting. 
For example, in offline fair division, Pareto efficiency and envy-freeness ex ante
are possible to achieve simultaneously, e.g.\ the allocations returned
by the probabilistic serial mechanism are Pareto efficient and
envy-free ex ante \cite{probserial}. In online fair division, these two normative properties are also possible to achieve simultaneously with 0/1 utilities, For instance, the {\sc Like} mechanism, allocating each next item uniformly at random to any agent who declares strictly positive utility, is Pareto
efficient and envy-free ex ante. 
But, with general (i.e.\ non-0/1) utilities, there is no
online mechanism that can be both Pareto efficient
and envy-free ex ante (proved first in \cite{awpricai2019}). 

\begin{mytheorem}
With general utilities, no online mechanism 
is envy-free ex ante and Pareto efficient ex ante.
\end{mytheorem}
\myproof
Let us consider the below simple fair
division problem with two agents and two arriving items.

\begin{center}
\begin{tabular}{|c|c|c|} \hline
  agent & item 1 & item 2  \\ \hline
   1 & $1$ & $2$ \\
  2 & $2$ & $1$ \\ \hline
\end{tabular}
\end{center}

To ensure envy-freeness ex ante, a mechanism must give to each agent a
probability of 1/2 for each item. 
This is not Pareto efficient ex ante. Giving each agent just their
most valued item Pareto dominates. 
\myqed

\begin{figure*}[htb]
\centering
\resizebox{0.75\textwidth}{!}{
\includegraphics[width=1\textwidth]{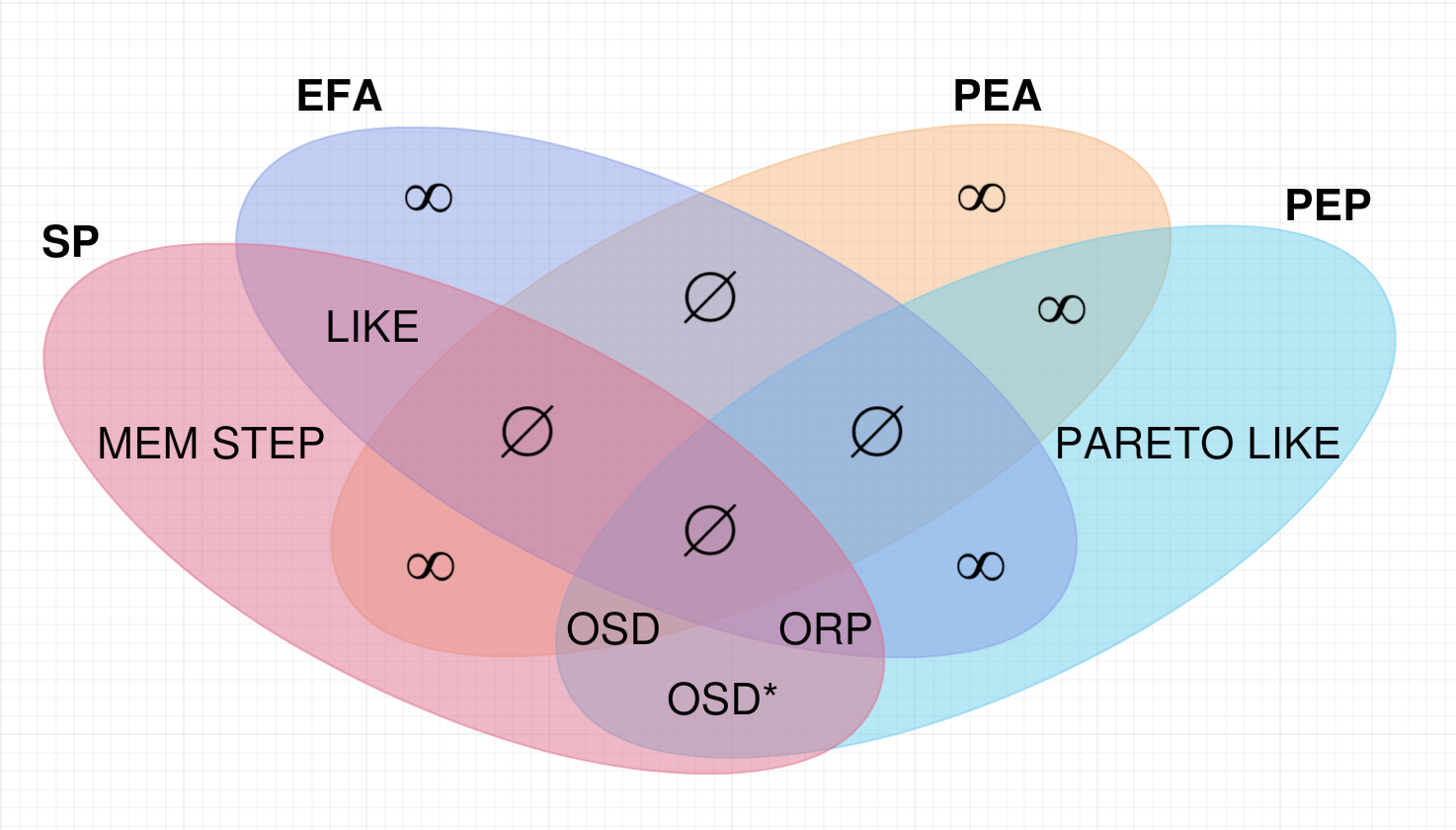}
}
\caption{Summary of results about online mechanisms 
satisfying (combinations of) properties such 
as strategy proofness (SP), 
envy-freeness ex ante (EFA), 
Pareto efficiency ex ante (PEA) and Pareto efficiency ex post
(PEP). Any part of the Venn diagram marked ``$\emptyset$'' is
impossible (e.g. no mechanism is envy-free
ex ante and Pareto efficient ex post). 
Any part of the Venn diagram marked ``$\infty$'' 
represents a combination of properties
which an infinite number of different mechanisms
can satisfy (e.g. any probabilistic combination of the
{\sc Like} and {\sc Orp} mechanisms returns a
probability distribution of allocations that
is envy-free ex ante).
{\sc Osd*} represents any probability distribution of 
online serial dictator ({\sc Osd}) mechanisms. {\sc Mem Step} is any ``memoryless step''
mechanism where the current item is allocated 
irrespective of the past (i.e.\ memoryless), and
whether an agent is feasible is simply a step
function of their bid. 
This figure is taken from \cite{awpricai2019}. 
}
\label{fig:one}
\end{figure*}

In Figure~\ref{fig:one}, we summarize a range of possibility and impossibility
results that characterize whether there exist mechanisms that
have combinations of properties such as 
strategy proofness (SP), envy-freeness ex ante (EFA), 
Pareto efficiency ex ante (PEA) and Pareto efficiency ex post
(PEP).

\subsection{Intractability results}

Another angle to understanding online fair
division is to study the complexity of 
relevant computational problems such as
computing outcomes or strategic actions. 
For example, consider again the 
{\sc Like} mechanism and the related
{\sc Balanced Like} mechanism which 
allocates the next item uniformly at random to any agent
who declares strictly positive utility for the item 
from amongst the agents currently allocated the fewest items \cite{aagwijcai2015}. 
The 
{\sc Balanced Like} mechanism is designed
to balance the number of items allocated
to each agent. 

Now, the {\sc Like} mechanism is strategy-proof. If
an agent has strictly positive utility for an 
item, they should bid for it. 
On the other hand, the 
{\sc Balanced Like} mechanism is not strategy-proof,
even when restricted to 0/1 utilities. Agents may decide not to bid now, even when they have strictly positive utility for an arriving item, as it may be more profitable to wait for an item that arrives later which is less competitive. Furthermore, they may thus receive multiple items in future. 

In response, Aleksandrov and Walsh (\cite{awki17}) consider the computational
complexity for an agent to compute strategic bids
to improve their (expected) outcome. 
Such computational problems are 
intractable in general. So the fact that the {\sc Balanced Like} mechanism is not strategy-proof is tempered by the intractability of computing strategic actions. In more general terms, there is a trade-off between computability of strategic actions and achievability of fairer allocations.

\subsection{Asymptotic guarantees}

Another response to the challenge of online fair division is to
try to achieve properties asymptotically. Even in the offline setting,
desirable fairness property such as envy-freeness cannot be guaranteed. Consider, 
for instance, two agents and one indivisible item that both agents
like. Envy-freeness is even harder to guarantee in the online setting.
Suppose two agents both like the next indivisible item. If we want to 
ensure envy-freeness, we cannot allocate this item fairly
without ultimately knowing what items will arrive next,
that might compensate the agent not allocated the current
item. 

However, we can look for online mechanisms that limit how envy
grows over time. For example, \cite{procacciaonlineenvy} consider
whether mechanisms can achieve {\em vanishing envy}.  That is,
after $m$ items have arrived in an online fashion, if the maximum amount
of envy an agent has for any other agent is $envy_m$, can we 
design mechanisms so that the ratio of $\frac{envy_m}{m}$
goes to zero as $m$ goes to infinity?
This is indeed possible and, in fact, easy to achieve. 
\cite{procacciaonlineenvy}  show that randomly allocating items 
gives envy that vanishes in expectation. Unfortunately,
it is not possible to do better than such a ``blind'' mechanism. Indeed, the ran- dom mechanism is asymptotically optimal up to logarithmic factors.

\subsection{Reallocating items}

A final technique to deal with the online nature of fair division
is to consider allocations that can be adjusted once
subsequent items are revealed. By re-allocating items
allocated in the past when new items arrive, we can
perhaps restore desirable normative properties. 
For example, with two agents, additive valuations and $m$ online items, 
\cite{procacciaonlinepast} prove that any
uninformed algorithm requires $\Theta(m)$ items to be 
re-allocated to ensure an EF1 allocation. Unsurprisingly, 
an informed algorithm requires no re-allocations to ensure EF1. For the case of three or more agents, 
\cite{procacciaonlinepast} 
prove that even informed algorithms require
$\Omega(m)$ items to be reallocated to ensure an EF1 allocation, and design 
an uninformed algorithm that makes do with $O(m^{3/2})$. 
These results leave open the question of whether there is a separation between the number of items that informed and uninformed algorithms must re-allocate in order to achieve EF1 (like the strict separation proven in the case of two agents) in the general case of three or more agents.

\section{Conclusions}

Online fair division is a promising and active research area
that looks to take account of the online nature 
of many resource allocation problems. 
Whilst recent work has identified many 
important features of such problems, 
there are many research directions still
to be considered. For example, 
how do we extend recent results such as 
\cite{agmwaij2015,agmmnwaamas15,aswijcai16,arswaaai2017,gdagmmwijcai2019,aciwijcai2019,awki18} about
the (offline)
fair division of goods and bads to the online
setting? As a second example, 
the preferences of agents are often
highly correlated. How can we exploit
this fact in an online setting?
As a third example, we can apply 
online mechanisms like those discussed here 
to an offline setting
by imposing an ``artificial'' arrival order to the agents or items. 
Can this help achieve desirable normative properties?
And, as a fourth example, can we extend
these ideas to the online versions of other problems in
social choice such as capacitated facility location \cite{acllwaaai2020} or
peer assessment \cite{wecai2014}.

\section{Acknowledgments}

The authors are funded by the European Research Council 
under the Horizon 2020 Programme via an Advanced 
Research Grant AMPLify 670077. 

\bibliographystyle{aaai}
\bibliography{a-z2,pub2}

\end{document}